\documentclass{article}
\usepackage{arxiv_template}
\newcommand{\gray}[1]{{\color{gray} #1}}

\title{\fontsize{13pt}{16pt}\selectfont%
MinkowskiPE: Minkowski Positional Encoding for Spatiotemporal Perception
}

\author{%
Yuhao Li$^{1,2,*}$, ~ 
Louie Hong Yao$^{1}$, ~
Tianyi Shi$^{1,2}$, ~
Hanqun Cao$^{2}$, \\
Hongxia Hao$^{3}$, ~
Zhen Zhao$^{3}$, ~
Shengchao Liu$^{1,2,\dagger}$ \\[0.5em]
$^{1}$ \textit{Wave Intelligence Lab} \\ 
$^{2}$ \textit{Department of Computer Science and Engineering, The Chinese University of Hong Kong} \\
$^{3}$ \textit{Shanghai Artificial Intelligence Laboratory} \\[0.5em]
$^{*}$ liyh5366@gmail.com \quad $^{\dagger}$ scliu@cuhk.edu.hk
}

\date{\today}

\begin{document}

\maketitle

\begin{abstract} \vspace{-0.5em}
Modeling spatiotemporal coupling is a key challenge in building physical intelligence across scales, from microscopic to macroscopic. 
Existing models capture such structure broadly through physics-motivated dynamical formulations or learning-motivated architectures. The former provide stronger priors but may constrain flexibility, whereas the latter are more flexible but leave the spatiotemporal coupling largely implicit. We therefore seek an approach that combines flexible learning with an explicit geometric bias for jointly modeling time and space.
To this end, we propose \emph{Minkowski Positional Encoding} (MinkowskiPE), which uses joint temporal and spatial coordinates to parameterize Lorentz transformations applied to query and key features. With MinkowskiPE, the query–key attention score depends on position only through the relative spacetime displacement between the two tokens and is therefore invariant to global translation of the coordinates. This paradigm retains the standard dot-product attention interface and remains compatible with efficient attention implementations. 
We evaluate MinkowskiPE on microscopic molecular dynamics and macroscopic video prediction tasks, achieving the best results on all nine multi-trajectory molecular evaluations and reducing KTH video-prediction MSE by 9.9\% relative to the best baseline while using roughly one-tenth as many parameters.
\end{abstract}

\section{Introduction}
\label{sec:intro}

Space and time are fundamental to the description of physical systems across scales, from microscopic molecular motion to macroscopic scene dynamics. Accordingly, physical intelligence requires the ability to capture their joint spatiotemporal structure. In molecular dynamics, each atomic state is associated with a time step and a position in three-dimensional space \cite{MISATO, NeuralMD}. In video, each image patch is associated with a frame and a spatial location within that frame \cite{TimeSformer, ViViT}. Despite their differences in scale, modality, and data structure, both settings can therefore be viewed as collections of observations with spatiotemporal coordinates.

\begin{figure}[t]
  \centering
  \includegraphics[width=\textwidth]{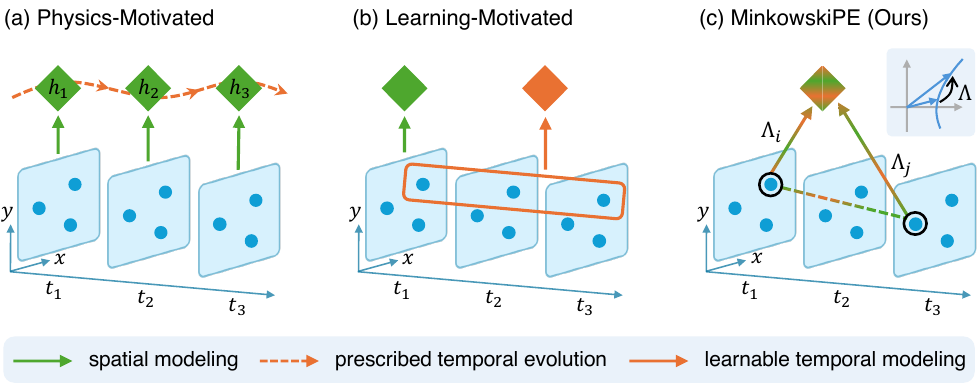}
  \vspace{-2ex}
  \caption{%
  Different perspectives on spatiotemporal modeling.
  (a) Physics-motivated approaches impose explicit structure on temporal evolution. 
  (b) Learning-motivated approaches learn spatiotemporal dependencies directly from data. 
  (c) MinkowskiPE combines physics-inspired geometric priors with flexible data-driven learning.
  }
  \label{fig:pipeline_comparison}
\end{figure}

Existing methods for spatiotemporal modeling broadly fall into two categories. The first research line is motivated by the view of temporal modeling as dynamical evolution, as in \Cref{fig:pipeline_comparison}(a). This includes recurrent and convolutional propagation \cite{ConvLSTM, PredRNN, SimVP}, as well as continuous-time evolution parameterized by differential equations \cite{NeuralODE, NeuralMD}. Other models incorporate stronger dynamical structure through physical priors or constraints \cite{PINN, HNN, MeshGraphNet}. Across these methods, temporal structure is organized through a prescribed propagation or evolution mechanism, introducing stronger inductive structure but potentially limiting the model's flexibility in learning spatiotemporal dependencies from data.

The second research line is learning-motivated, focusing on the data-driven modeling of spatiotemporal coupling, as in \Cref{fig:pipeline_comparison}(b). Transformer-based models usually capture the coupling of time and space through the structure of learned attention, such as factorized spatial and temporal attention, multiscale attention, and trajectory-aware attention \cite{TimeSformer, ViViT, MViT, Motionformer}. Such models provide more flexibility in modeling how information interacts across space and time, but the relation between temporal and spatial coordinates generally remains implicit.

These two directions highlight a trade-off between physical structure and data-driven flexibility in spatiotemporal coupling: physics-motivated methods often limit expressiveness, whereas learning-based methods often lack explicit physical priors. This naturally raises a question: \emph{Can we design a spatiotemporal coupling paradigm that combines the flexibility of data-driven learning with a strong physical prior?} To explore this question, we focus on positional encoding as a mechanism for modeling temporal and spatial coordinates while preserving their distinct roles \cite{Attention}. Further, Minkowski geometry provides such a structure, placing time and space within a unified coordinate system while assigning them different geometric roles \cite{Carroll2019}.

\textbf{Our Contributions.} We propose \emph{Minkowski Positional Encoding} (MinkowskiPE), a relative positional encoding for spatiotemporal modeling, as illustrated in~\Cref{fig:pipeline_comparison}(c). Each token is associated with a spacetime coordinate that parameterizes Lorentz transformations applied to its query and key features. For multidimensional spacetime coordinates, we learn multiple projection directions and apply the corresponding transformations across feature blocks. As a result, the pairwise attention depends on position only through the relative spacetime displacement. MinkowskiPE preserves the standard dot-product attention interface, allowing it to directly benefit from efficient attention implementations such as FlashAttention \cite{FlashAttention}.

Empirically, we evaluate MinkowskiPE on two spatiotemporal prediction tasks, molecular dynamics and video prediction. Molecular dynamics involves the three-dimensional motion of atoms over time, whereas video prediction concerns the evolution of two-dimensional visual scenes across frames. Together, these tasks evaluate the same positional formulation across different physical scales, data modalities, and spatial structures. On molecular dynamics prediction, MinkowskiPE ranks among the top two in 26 of 30 evaluations across ten protein--ligand systems in the single-trajectory setting, and achieves the best performance on all nine evaluations across three multi-trajectory datasets. On KTH video prediction, MinkowskiPE achieves the best performance on four of five metrics among the compared methods, reducing MSE by 9.9\% while using roughly one-tenth as many parameters as the lowest-MSE baseline.

\section{Related Work}
\label{sec:related-work}

\textbf{Spatiotemporal Modeling.} Existing models encode temporal evolution through a range of architectural and dynamical mechanisms. Recurrent and convolutional models organize temporal information through sequential propagation or local temporal transformations \cite{ConvLSTM, PredRNN, SimVP}, while continuous-time approaches parameterize evolution using differential equations \cite{NeuralODE, NeuralMD}. Other methods incorporate stronger dynamical or physical structure through differential-equation constraints \cite{PINN, PhyDNet}, Hamiltonian dynamics \cite{HNN}, or graph-based simulation \cite{GNS, MeshGraphNet}. Transformer-based models instead learn spatiotemporal dependencies through attention mechanisms such as factorized spatial--temporal attention \cite{TimeSformer, ViViT}, multiscale representations \cite{MViT}, local windows \cite{VideoSwin, Earthformer}, and trajectory-aware attention \cite{Motionformer}. Our work differs by introducing inductive bias at the positional representation level rather than prescribing temporal evolution or a specific attention organization.

\textbf{Positional Representations.}
Transformers incorporate positional information through learned relative embeddings \cite{RelativePE}, relative attention formulations \cite{TransformerXL}, attention biases \cite{T5-RPB, ALiBi}, and feature-space transformations \cite{RoPE}. These mechanisms were later generalized to multidimensional coordinates in vision and multimodal models \cite{RoPE-ViT, FiT, VisionLLaMA, Qwen2-VL}, and further developed through Lie-group transformations, commuting operators, translation-invariant constructions, and group representations \cite{LieRE, ComRoPE, STRING, RethinkingRoPE, GRAPE, GeoPE}. Related work has also explored hyperbolic feature transformations for one-dimensional positional encoding \cite{HoPE}. More recent work has extended positional representations to spatiotemporal settings, such as video \cite{CogVideoX, VideoRoPE, VRoPE}. MinkowskiPE instead uses joint temporal and spatial event coordinates to parameterize Lorentz transformations within a single relative positional mechanism.

\section{Preliminaries}
\label{sec:prelim}

\subsection{Relative Positional Encoding}
\label{sec:prelim_rpe}

Let $q_i,k_j \in \mathbb{R}^{d_\text{em}}$ denote the query and key vectors associated with positions $x_i$ and $x_j$. A relative positional encoding is designed such that their pairwise interaction depends on the positions only through their relative displacement $\Delta x_{ij} = x_j - x_i$, \ie{},
\begin{equation} \label{eq:relative_pe}
    \left\langle
    f_q (q_i, x_i), \, f_k(k_j, x_j)
    \right\rangle
    = g(q_i, k_j, \Delta x_{ij}),
\end{equation}
where $\left\langle \cdot, \cdot \right\rangle$ denotes the dot product in feature space, $f_q$ and $f_k$ denote feature transformations, and $g$ denotes the resulting query--key score before attention scaling and softmax.

Equivalently, this score is invariant under a common translation $x_i \mapsto x_i + a$ and $x_j \mapsto x_j + a$. Existing positional encodings realize this property through relative embeddings, attention biases, or position-dependent transformations of query and key features \citep{RelativePE, T5-RPB, ALiBi, RoPE}. 

\subsection{Minkowski Geometry and Lorentz Transformations}
\label{sec:prelim_minkowski}

We next introduce the geometric and algebraic structure used in our construction. A spatiotemporal \emph{event} in $(1+d)$-dimensional spacetime is represented by
\begin{equation} \label{eq:event_coordinate}
    x = (t, r) \in \mathbb{R}^{1+d},
\end{equation}
where $t \in \mathbb{R}$ denotes the temporal coordinate and $r \in \mathbb{R}^d$ is the spatial coordinate.

Minkowski geometry distinguishes temporal and spatial directions through an indefinite \emph{metric}
\begin{equation} \label{eq:Minkowski_metric}
    \eta = \mathrm{diag} (-1, +1, \cdots, +1) \in \mathbb{R}^{(1+d) \times (1+d)}.
\end{equation}
The metric $\eta$ defines the indefinite bilinear form $u^\top \eta v$ on spacetime coordinates. For two events $x_i$ and $x_j$, the relative displacement $\Delta x_{ij}=x_j-x_i$ has the associated quadratic form
\begin{equation} \label{eq:minkowski_interval}
    \Delta x_{ij}^\top \eta \Delta x_{ij}
    = - (t_j -t_i)^2 + \| r_j - r_i \|^2 .
\end{equation}
The signature of $\eta$ distinguishes the temporal direction from the spatial directions, causing them to enter this form with opposite signs. The transformations preserving this bilinear form are \emph{Lorentz transformations}. A matrix $\Lambda$ is Lorentz if
\begin{equation} \label{eq:lorentz_condition}
    \Lambda^\top \eta \Lambda = \eta,
\end{equation}
or equivalently,
\begin{equation} \label{eq:lorentz_inverse}
    \Lambda^{\top} \eta = \eta \Lambda^{-1}.
\end{equation}
This identity will be the key algebraic ingredient for constructing a relative positional interaction.

Notice that in the proposed MinkowskiPE method, the event coordinates themselves are not transformed by $\Lambda$. Instead, they parameterize Lorentz transformations applied to query--key features. We therefore use Minkowski geometry as a positional-representation structure rather than imposing Lorentz symmetry on the underlying data.

\section{Method}
\label{sec:method}

Building on the relative position construction in \Cref{sec:prelim}, we formulate spatiotemporal tokens as events and construct a Lorentzian positional transformation whose pairwise interaction depends only on relative event displacement. Then we extend this construction to practical multi-head attention.

\textbf{Event-Token Formulation.}
We consider a sequence of tokens $\{(h_i, x_i)\}_{i=1}^N$, where $h_i \in \mathbb{R}^{d_{\mathrm{em}}}$ denotes the embedding of the token $i$, and $x_i$ denotes the associated spatiotemporal coordinate. We write the coordinate as $x_i = (t_i, r_i)$, with $t_i \in \mathbb{R}$ being the temporal coordinate, $r_i \in \mathbb{R}^{d}$ being the spatial coordinate. 
For two events, we define their relative displacement as $\Delta x_{ij} = x_j - x_i$. This event-based representation applies to a broad range of spatiotemporal applications. For example, in video, a token may correspond to an image patch at a particular frame, with $r_i \in \mathbb{R}^{2}$. In molecular dynamics, a token may correspond to an atom at a particular time, with $r_i \in \mathbb{R}^{3}$.

\subsection{Minkowski Positional Encoding} 
\label{sec:method_pe}

We seek to use the spatiotemporal coordinate $x_i$ to parameterize a Lorentz transformation acting on query and key features. Following \Cref{eq:lorentz_inverse}, we have
\begin{equation} \label{eq:lorentz_pairwise}
    \Lambda(x_i)^\top \eta \Lambda(x_j) 
    = \eta \Lambda(x_i)^{-1}\Lambda(x_j).
\end{equation}
To ensure that the positional interaction depends only on relative displacement $\Delta x_{ij} = x_j - x_i$, a sufficient condition is
\begin{equation} \label{eq:homomorphism}
    \Lambda(x_1 + x_2) = \Lambda(x_1) \Lambda(x_2),
    \quad \Lambda(0) = I,
\end{equation}
which gives $\Lambda(x_i)^{-1}\Lambda(x_j) = \Lambda(x_j-x_i)$.

For multidimensional coordinates, a general Lorentz representation satisfying \Cref{eq:homomorphism} requires mutually commuting transformations across coordinate directions. To obtain a simple and tractable construction, we restrict it to a \emph{one-parameter} Lorentz subgroup. 
The event coordinate $x_i \in \mathbb{R}^{D}$, with $D=1+d$, is first mapped to a scalar parameter
\begin{equation} \label{eq:projection_vector}
    \rho_i = w^\top x_i,
\end{equation}
by a learnable linear projection $w \in \mathbb{R}^{D}$, so that $\rho_j - \rho_i = w^\top \Delta x_{ij}$.
We therefore parameterize the transformation associated with event $x_i$ by $\Lambda(\rho_i)$.
Appendix~\ref{app:derivation} gives a group-theoretic characterization of the construction, showing that the homomorphic formulation above imposes commuting-generator structure and that, for 2D Lorentzian feature blocks, any smooth homomorphism from the additive group into $SO^+(1,1)$ reduces to the scalar-projection one-parameter form used above.

\paragraph{MinkowskiPE.} 
We then instantiate the one-parameter construction above using the standard (1+1)-dimensional Lorentz boost acting on two-dimensional query and key feature blocks
\begin{equation} \label{eq:lorentz_boost}
    \Lambda(\rho) =
    \begin{bmatrix}
        \cosh\rho & \sinh\rho\\
        \sinh\rho & \cosh\rho
    \end{bmatrix}, \quad
    \eta =
    \begin{bmatrix}
        -1 & 0\\
        0 & 1
    \end{bmatrix}.
\end{equation}
Let $q_i, k_i \in \mathbb{R}^{2}$ denote a query and key feature block associated with event $x_i$, with $\rho_i=w^\top x_i$ defined as above. 
We define the position-encoded query and key as
\begin{equation} \label{eq:qk_transformation}
    \tilde q_i = \Lambda(\rho_i)q_i,
    \quad
    \tilde k_i = \eta\Lambda(\rho_i)k_i.
\end{equation}
The metric factor $\eta$ in the key transformation allows the standard Euclidean dot product to realize the corresponding Minkowski bilinear interaction. Specifically,
\begin{equation} \label{eq:inner_product}
    \tilde q_i^\top \tilde k_j 
    = q_i^\top \Lambda(\rho_i)^\top \eta \Lambda(\rho_j) k_j
    = q_i^\top \eta \Lambda(\rho_j-\rho_i) k_j
    = q_i^\top \eta \Lambda\!\left(w^\top\Delta x_{ij}\right) k_j.
\end{equation}
Thus, although each token is transformed according to its own event coordinate, the positional contribution to the pairwise query--key interaction depends on $x_i$ and $x_j$ only through their relative displacement $\Delta x_{ij}$, as shown in \Cref{eq:inner_product}.

\subsection{Practical Parameterization}

The single-block construction in \Cref{sec:method_pe} extends directly to a full attention head by assigning a separate projection direction to each two-dimensional query--key feature block. For a head dimension $d_h$, we partition the query and key features into $P=d_h/2$ two-dimensional blocks and introduce a learnable projection matrix 
\begin{equation} \label{eq:projection_matrix}
    W = [w_1, \ldots, w_P] \in \mathbb{R}^{D\times P},
\end{equation}
where each column $w_p\in\mathbb{R}^D$ defines the positional projection for the $p$-th block. The same projection matrix $W$ is shared across all Transformer layers and attention heads.

For coordinate $x_i$, this produces the block-wise parameters $\rho_i=W^\top x_i \in \mathbb{R}^{P}$, with $\rho_{i,p}=w_p^\top x_i$. By linearity, $\rho_{j,p}-\rho_{i,p}=w_p^\top\Delta x_{ij}$, so each block receives a different learned scalar projection of the same relative event displacement.

For efficient implementation, the block-wise Lorentz transformations can be written equivalently in element-wise form. 
For $q_i=(q_{i,1}, \ldots, q_{i,d_h})^\top$, we have
\begin{equation} \label{eq:efficient_lorentz}
    \tilde q_i
    =
    \begin{bmatrix}
    q_{i,1}\\ q_{i,2}\\ q_{i,3}\\ q_{i,4}\\
    \vdots\\ q_{i,d_h-1}\\ q_{i,d_h}
    \end{bmatrix}
    \odot
    \begin{bmatrix}
    \cosh \rho_{i,1}\\ 
    \cosh \rho_{i,1}\\
    \cosh \rho_{i,2}\\
    \cosh \rho_{i,2}\\
    \vdots\\
    \cosh \rho_{i,P}\\
    \cosh \rho_{i,P}
    \end{bmatrix}
    +
    \begin{bmatrix}
    q_{i,2}\\ q_{i,1}\\ q_{i,4}\\ q_{i,3}\\
    \vdots\\ q_{i,d_h}\\ q_{i,d_h-1}
    \end{bmatrix}
    \odot
    \begin{bmatrix}
    \sinh \rho_{i,1}\\
    \sinh \rho_{i,1}\\
    \sinh \rho_{i,2}\\
    \sinh \rho_{i,2}\\
    \vdots\\
    \sinh \rho_{i,P}\\
    \sinh \rho_{i,P}
    \end{bmatrix},
\end{equation}
where $\odot$ denotes element-wise multiplication. The key features are transformed analogously, together with the Minkowski metric factor defined in \Cref{sec:method_pe}.

\begin{figure}[t]
    \centering
    \includegraphics[width=\linewidth]{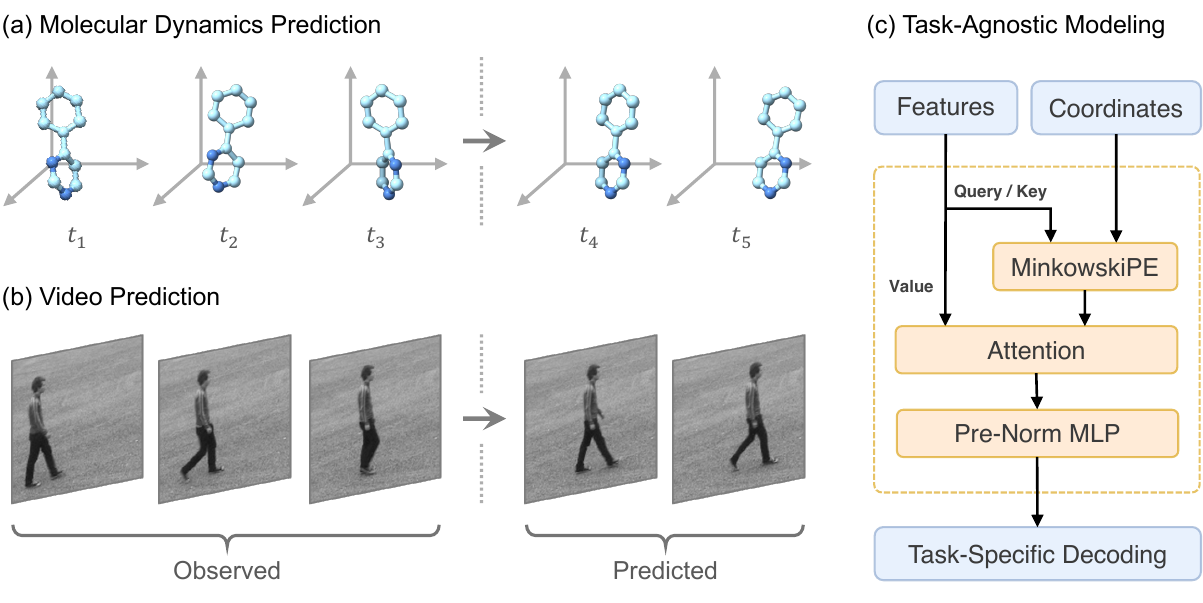}
    \vspace{-3ex}
    \caption{%
    Experimental overview.
    (a) Molecular dynamics and (b) video predictions instantiate spatiotemporal forecasting at different physical scales.
    (c) Both tasks follow the same task-agnostic modeling scheme, in which task-specific features are paired with spatiotemporal coordinates, processed by a Transformer with MinkowskiPE, and mapped to future states by task-specific decoders.
    }
    \label{fig:exp}
\end{figure}

\section{Experiments}
\label{sec:exp}

As illustrated in \Cref{fig:exp}, we evaluate MinkowskiPE on molecular dynamics and video prediction, two spatiotemporal prediction tasks that differ in physical scale and data structure. Molecular dynamics represents atomic motion in three-dimensional space, whereas video prediction represents visual evolution on a two-dimensional image plane. Despite these differences, both domains can be modeled within the same task-agnostic spatiotemporal framework. We describe the task formulation, model instantiation, and experimental protocol for each domain below.

\subsection{Molecular Dynamics Prediction}
\label{sec:exp_md}

\textbf{Task Formulation.} 
We consider the protein--ligand binding dynamics prediction under the semi-flexible setting, where the protein is treated as a fixed environment while the ligand evolves dynamically over time. Given the surrounding protein structure and an observed segment of ligand motion, the task is to predict the future three-dimensional positions of all ligand heavy atoms over multiple time steps. This requires the model to capture both the temporal evolution of individual atoms and the spatial interactions that constrain the ligand trajectory within its molecular environment.

\textbf{Model Instantiation.} 
For molecular dynamics, each ligand atom at each observed time step is treated as an event token with spatiotemporal coordinate $(t,x,y,z)$, where $t$ denotes the time step and $(x,y,z)$ specifies its three-dimensional position. The token features encode molecular identity and motion information, while the protein is represented as static residue-level structural context. Each Transformer block first performs frame-wise cross-attention from ligand atoms to protein residues, followed by global self-attention over ligand event tokens across the observed trajectory. MinkowskiPE is applied within the attention modules using the associated spatiotemporal coordinates. After the final Transformer block, the representation of each ligand atom at the last observed frame is passed to an MLP decoder, which predicts the displacements over all future frames.

\textbf{Experimental Setup.}
We evaluate on the MISATO dataset \citep{MISATO} under both single-trajectory and multi-trajectory settings. For single-trajectory prediction, a separate model is trained for each of the ten protein--ligand systems, using 10 observed frames to predict the subsequent 20 frames. For multi-trajectory prediction, a separate model is trained on each of MISATO-100, MISATO-1000, and MISATO-Full, which represent progressively larger collections of protein--ligand trajectories, using 80 observed frames to predict the subsequent 20 frames.

We evaluate prediction quality using MAE, Matching, and Stability, which measure coordinate accuracy and structural consistency from complementary perspectives. We compare against VerletMD, GNN-MD, and DenoisingLD, three representative baselines implemented in NeuralMD \citep{NeuralMD} and motivated by numerical integration, autoregressive graph-based simulation \citep{Fu2023CGMD}, and diffusion-based molecular dynamics \citep{Arts2023}, respectively. We further include NeuralMD-ODE, NeuralMD-SDE \citep{NeuralMD}, and the recent generative model BioMD \citep{BioMD}. More experimental details are provided in Appendix \ref{app:experiments}.

\begin{table}[H]
\centering
\caption{Results on single-trajectory predictions (in10/out20). All results are averaged over three random seeds. The best result in each column is shown in bold, and the second-best is underlined.}
\label{tab:md_single}
\begin{adjustbox}{width=\textwidth}
\begin{tabular}{llcccccccccc}
\toprule
Method & Metric & 5WIJ & 4ZX0 & 3EOV & 4K6W & 1KTI & 1XP6 & 4YUR & 4G3E & 6B7F & 3B9S \\
\midrule
\multirow{3}{*}{VerletMD} & MAE & 14.629 & 21.278 & 27.960 & 15.428 & 18.157 & 13.753 & 16.764 & 5.111 & 31.934 & 19.473 \\
 & Matching & 5.459 & 7.971 & 13.588 & 7.505 & 7.467 & 4.672 & 9.555 & 3.388 & 21.691 & 0.923 \\
 & Stability & 24.360 & 19.168 & 13.067 & 15.441 & 19.352 & 28.129 & 16.542 & 31.852 & 11.050 & 57.801 \\
\midrule
\multirow{3}{*}{GNN-MD} & MAE & 2.280 & 2.370 & 3.512 & 3.695 & \underline{6.641} & 2.378 & 7.031 & 2.709 & 4.136 & \underline{2.578} \\
 & Matching & 0.803 & 0.555 & 1.216 & 1.038 & 0.386 & 0.966 & 0.920 & 0.893 & 1.194 & 1.414 \\
 & Stability & 54.475 & 68.613 & 40.984 & 42.480 & 81.831 & 49.239 & 47.555 & 61.802 & 39.067 & 49.306 \\
\midrule
\multirow{3}{*}{NeuralMD-ODE} & MAE & \underline{2.252} & \textbf{1.878} & 3.858 & 3.656 & 6.675 & \underline{1.924} & 6.957 & \underline{2.191} & 3.921 & 3.039 \\
 & Matching & 0.464 & \underline{0.428} & 1.062 & 0.928 & 0.337 & \underline{0.537} & 0.584 & 0.505 & 0.459 & 0.659 \\
 & Stability & 82.046 & \underline{81.401} & 47.328 & 49.438 & 86.430 & 75.533 & 69.775 & 71.436 & 75.692 & 76.065 \\
\midrule
\multirow{3}{*}{NeuralMD-SDE} & MAE & 2.260 & 2.158 & \textbf{3.395} & 3.765 & 6.646 & 2.061 & 7.038 & 2.345 & \textbf{3.842} & 3.132 \\
 & Matching & 0.615 & 0.696 & \underline{0.962} & 1.076 & 0.167 & 0.615 & 0.749 & 0.521 & 0.741 & 0.444 \\
 & Stability & 67.464 & 59.109 & 50.108 & 49.700 & \underline{98.508} & 69.423 & 60.344 & 68.729 & 57.917 & \underline{77.801} \\
\midrule
\multirow{3}{*}{DenoisingLD} & MAE & 2.501 & 3.138 & 4.055 & 3.942 & 7.051 & 2.218 & 7.128 & 3.588 & 4.431 & 2.811 \\
 & Matching & 0.815 & 1.072 & 1.209 & 0.839 & 0.268 & 0.676 & 0.834 & 1.069 & 0.672 & 0.472 \\
 & Stability & 52.418 & 44.228 & 41.469 & 53.820 & 91.986 & 64.951 & 49.676 & 40.823 & 61.583 & 71.852 \\
\midrule
\multirow{3}{*}{BioMD} & MAE & 3.191 & 2.200 & 4.193 & \underline{3.362} & \textbf{4.251} & 2.273 & \textbf{5.163} & 2.535 & 5.156 & 4.243 \\
 & Matching & \textbf{0.386} & \textbf{0.424} & \textbf{0.712} & \underline{0.778} & \underline{0.153} & 0.544 & \underline{0.477} & \underline{0.228} & \textbf{0.189} & \underline{0.095} \\
 & Stability & \textbf{87.150} & \textbf{83.080} & \textbf{64.430} & \underline{57.680} & 98.320 & \underline{78.050} & \underline{78.840} & \underline{94.420} & \textbf{96.530} & \textbf{100.000} \\
\midrule
\multirow{3}{*}{MinkowskiPE} & MAE & \textbf{2.143} & \underline{1.982} & \underline{3.491} & \textbf{2.477} & 6.669 & \textbf{1.911} & \underline{6.491} & \textbf{1.818} & \underline{3.899} & \textbf{2.486} \\
 & Matching & \underline{0.392} & 0.441 & 0.989 & \textbf{0.742} & \textbf{0.129} & \textbf{0.509} & \textbf{0.422} & \textbf{0.190} & \underline{0.205} & \textbf{0.071} \\
 & Stability & \underline{86.882} & 81.138 & \underline{52.287} & \textbf{58.499} & \textbf{99.383} & \textbf{79.397} & \textbf{81.934} & \textbf{97.325} & \underline{95.825} & \textbf{100.000} \\
\bottomrule
\end{tabular}
\end{adjustbox}
\end{table}

\begin{table}[ht]
\centering
\caption{Results on multi-trajectory predictions (in80/out20). All results are averaged over three random seeds.  The best result in each column is shown in bold, and the second-best is underlined.}
\label{tab:md_multiple}
\begin{adjustbox}{width=\textwidth}
\begin{tabular}{lccccccccc}
\toprule
\multirow{2}{*}{Method}
& \multicolumn{3}{c}{MISATO-100}
& \multicolumn{3}{c}{MISATO-1000}
& \multicolumn{3}{c}{MISATO-Full} \\
\cmidrule(lr){2-4} \cmidrule(lr){5-7} \cmidrule(lr){8-10}
& MAE & Matching & Stability
& MAE & Matching & Stability
& MAE & Matching & Stability \\
\midrule
VerletMD
& 14.317 & 6.927 & 19.78
& 27.374 & 10.042 & 20.10
& 24.242 & 10.101 & 18.76 \\
GNN-MD
& 4.198 & 0.489 & 77.67
& 4.500 & 0.553 & 75.06
& 4.434 & 0.509 & 77.59 \\
NeuralMD-ODE
& 4.210 & 0.407 & 84.76
& 4.548 & 0.446 & 84.69
& 4.448 & 0.428 & 84.90 \\
NeuralMD-SDE
& 4.215 & 0.407 & 84.71
& 4.555 & 0.446 & 84.59
& 4.455 & 0.430 & 84.70 \\
DenoisingLD
& \underline{4.184} & \underline{0.396} & \underline{85.39}
& \underline{4.467} & \underline{0.430} & \underline{85.42}
& 4.400 & \underline{0.412} & \underline{85.64} \\
BioMD
& 4.747 & 0.471 & 80.28
& 4.522 & 0.472 & 82.47
& \underline{4.390} & 0.420 & 85.15 \\
\midrule
MinkowskiPE
& \textbf{4.181} & \textbf{0.395} & \textbf{85.45}
& \textbf{4.367} & \textbf{0.415} & \textbf{86.01}
& \textbf{4.315} & \textbf{0.392} & \textbf{86.29} \\
\bottomrule
\end{tabular}
\end{adjustbox}
\end{table}

\begin{figure}[ht]
    \centering
    \includegraphics[width=\linewidth]{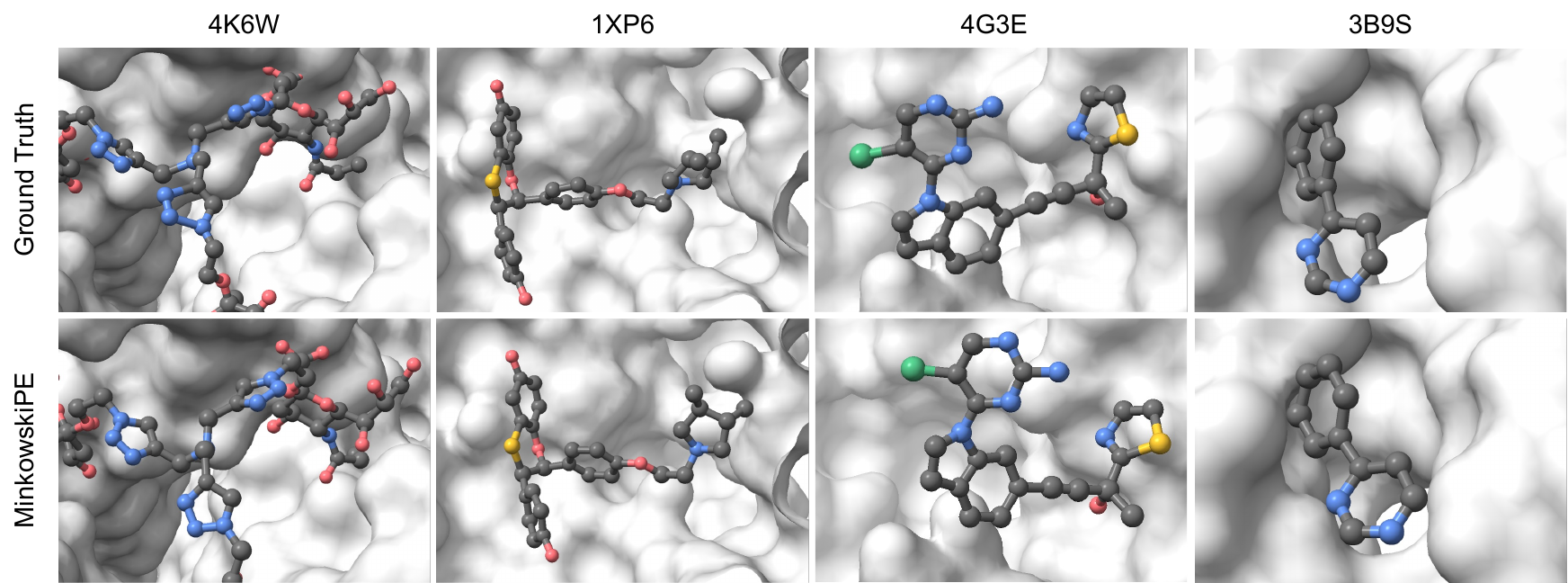}
    \vspace{-3ex}
    \caption{Representative molecular dynamics predictions. Final-step ligand configurations predicted by MinkowskiPE for 4K6W, 1XP6, 4G3E, and 3B9S, compared with the corresponding reference configurations. The surrounding protein surface is locally clipped for visualization clarity.}
    \label{fig:md_visualization}
\end{figure}

\textbf{Results.} 
\Cref{tab:md_single} reports the single-trajectory results.
Across the ten protein--ligand systems and three evaluation metrics, MinkowskiPE achieves the best result in 17 of 30 evaluations and ranks among the top two in 26 of 30. In particular, it achieves the best performance across all three metrics on 4K6W, 1XP6, 4G3E, and 3B9S. Its strong performance on both Matching and Stability indicates that the gains are not confined to coordinate accuracy, but also extend to the structural quality of the predicted ligand configurations. \Cref{fig:md_visualization} visualizes the final predicted configurations for these four systems alongside the corresponding ground truth, providing a direct view of the resulting molecular structures.

\Cref{tab:md_multiple} reports the multi-trajectory results.
Across all three dataset scales, MinkowskiPE achieves the best performance on all nine evaluations. The advantage is consistent across both coordinate and structural metrics and persists as the number and diversity of training trajectories increase. Together with the single-trajectory results, these findings show that MinkowskiPE performs well not only when modeling individual molecular systems, but also when learning shared dynamics across heterogeneous protein--ligand complexes.

\subsection{Video Prediction}
\label{sec:exp_video}

\textbf{Task Formulation.}
We next consider video prediction, where a model observes a sequence of past frames and forecasts the subsequent visual evolution. Unlike molecular dynamics, where spatial coordinates correspond to continuous three-dimensional atom positions, video observations are organized on a regular two-dimensional image plane. The task therefore requires modeling both temporal changes across frames and spatial structure within each frame, including the motion and deformation of visual patterns over time.

\textbf{Model Instantiation.} 
Each video frame is first encoded by a convolutional encoder into a $16\times16$ latent feature grid. Every latent feature is treated as an event token with spatiotemporal coordinate $(t,u,v)$, where $t$ denotes the frame index and $(u,v)$ specifies its spatial location on the latent grid. The observed tokens are concatenated with future tokens initialized from the last observed latent map, and augmented with learnable future-step embeddings. All tokens are then jointly processed by an eight-layer global Transformer equipped with MinkowskiPE. The resulting future latent features are reshaped into spatial feature maps and passed through a convolutional decoder to reconstruct all future frames.

\textbf{Experimental Setup.}
We evaluate on the KTH human-action dataset following the OpenSTL protocol \citep{OpenSTL}. The model takes 10 grayscale frames at $128\times128$ resolution as input and predicts the subsequent 20 frames. We evaluate prediction quality using MSE, MAE, SSIM, PSNR, and LPIPS, covering pixel-level accuracy, structural similarity, and perceptual quality. 

We compare against representative video-prediction baselines reported in OpenSTL, including ConvLSTM \citep{ConvLSTM}, MIM \citep{MIM}, PredRNN variants \citep{PredRNN, PredRNN++, PredRNNV2}, SimVP variants \citep{SimVP, SimVPv2}, and TAU \citep{TAU}. To match its reporting protocol, which selects the best model over three trials, we use our lowest-MSE run among three random seeds for the primary comparison with OpenSTL baselines. We additionally report the mean performance across the three seeds for completeness. More experimental details are provided in Appendix \ref{app:experiments}.

\begin{table}[t]
\small
\centering
\caption{Results on KTH video prediction (in10/out20). 
We adopt the baseline results and experimental settings from OpenSTL \citep{OpenSTL}, which report the best result across three random seeds. For protocol-matched comparison, bold and underline denote the best and second-best results using our best run. The three-seed mean of MinkowskiPE is additionally reported for reference.
}
\label{tab:video}
\vspace{-1ex}
\begin{adjustbox}{max width=\textwidth}
\begin{tabular}{lcccccc}
\toprule
Method & Params (M) & MSE $\downarrow$ & MAE $\downarrow$ & SSIM $\uparrow$ & PSNR $\uparrow$ & LPIPS $\downarrow$ \\
\midrule
ConvLSTM-S & 14.9 & 47.65 & 445.5 & 0.8977 & 26.99 & 0.26686 \\
MIM & 39.8 & 40.73 & 380.8 & 0.9025 & 27.78 & \underline{0.18808} \\
PredRNN & 23.6 & 41.07 & 380.6 & 0.9097 & 27.95 & 0.21892 \\
PredRNN++ & 38.3 & 39.84 & 370.4 & \textbf{0.9124} & \underline{28.13} & 0.19871 \\
PredRNN.V2 & 23.6 & \underline{39.57} & \underline{368.8} & \underline{0.9099} & 28.01 & 0.21478 \\
SimVP+IncepU & 12.2 & 41.11 & 397.1 & 0.9065 & 27.46 & 0.26496 \\
SimVP+gSTA & 15.6 & 45.02 & 417.8 & 0.9049 & 27.04 & 0.25240 \\
TAU & 15.0 & 45.32 & 421.7 & 0.9086 & 27.10 & 0.22856 \\
\midrule
MinkowskiPE (best) & 2.5 & \textbf{35.66} & \textbf{347.7} & 0.9078 & \textbf{28.18} & \textbf{0.18490} \\
\gray{MinkowskiPE (mean)} 
& \gray{2.5} & \gray{35.78} & \gray{347.9} & \gray{0.9079} & \gray{28.15} & \gray{0.18342}
\\
\bottomrule
\end{tabular}
\end{adjustbox}
\end{table}

\textbf{Results.}
\Cref{tab:video} reports the results on the KTH video prediction task. Under the protocol-matched comparison, MinkowskiPE achieves the best performance on MSE, MAE, PSNR, and LPIPS, while remaining competitive on SSIM. Compared with the strongest baseline result for each metric, it reduces MSE, MAE, and LPIPS by 9.9\%, 5.7\%, and 1.7\%, respectively, while using only 2.5M parameters. The three-seed mean remains better than the reported baseline results on the same four metrics. Overall, these results show that the same event-based positional formulation used for molecular dynamics also performs effectively on dense visual prediction with a different spatial dimensionality and token structure.

\subsection{Effect of MinkowskiPE}

\begin{table}[b]
\small
\centering
\caption{
Controlled ablation of MinkowskiPE on molecular dynamics and video prediction. We compare the same model architecture with MinkowskiPE and with positional encoding removed, while keeping all others unchanged. All results are averaged over three random seeds.
}
\vspace{-1ex}
\label{tab:ablation}
\begin{adjustbox}{max width=\textwidth}
\begin{tabular}{lcccccccc}
\toprule
\multirow{2}{*}{\raisebox{-0.6ex}{\textbf{Method}}}
& \multicolumn{3}{c}{\textbf{MISATO-1000}}
& \multicolumn{5}{c}{\textbf{KTH}} \\
\cmidrule(lr){2-4}
\cmidrule(lr){5-9}
& MAE $\downarrow$
& Matching $\downarrow$
& Stability $\uparrow$
& MSE $\downarrow$
& MAE $\downarrow$
& SSIM $\uparrow$
& PSNR $\uparrow$
& LPIPS $\downarrow$ \\
\midrule
No PE
& 4.488
& 0.432
& 84.96
& 37.93
& 360.8
& 0.9045
& 27.93
& 0.19203 \\
MinkowskiPE
& \textbf{4.367}
& \textbf{0.415}
& \textbf{86.01}
& \textbf{35.78}
& \textbf{347.9}
& \textbf{0.9079}
& \textbf{28.15}
& \textbf{0.18342} \\
\bottomrule
\end{tabular}
\end{adjustbox}
\end{table}

To isolate the contribution of the proposed positional mechanism from the underlying task-specific architectures, we conduct controlled ablations on both molecular dynamics and video prediction. In each case, we remove only the positional encoding from the corresponding attention modules, and keep the model architecture, optimization protocol, and training configuration unchanged. As shown in Table~\ref{tab:ablation}, MinkowskiPE improves all evaluated metrics in both domains. On MISATO-1000, it reduces MAE and Matching by 2.7\% and 3.9\% respectively, and improves Stability by 1.2\%. On KTH, it reduces MSE and MAE by 5.7\% and 3.6\% respectively, and improves LPIPS by 4.5\%, with consistent gains in SSIM and PSNR. These controlled comparisons show that the improvements extend across multiple evaluation metrics and two different spatiotemporal domains, supporting that the observed gains are not solely attributable to the task-specific architectures.

\section{Discussion} \label{sec:discussion}
\vspace{-1ex}

\textbf{Minkowski Structure as Representation Bias.}
MinkowskiPE does not assume that the underlying data obey relativistic dynamics. Rather, it uses the algebraic structure of Minkowski geometry as a representation-level inductive bias for spatiotemporal attention. Joint temporal and spatial coordinates parameterize Lorentz transformations in query--key feature space. Together with the linear coordinate projections, preservation of the indefinite bilinear form ensures that the positional interaction depends only on relative event displacement. This construction provides a geometric mechanism for jointly modeling temporal and spatial information without prescribing a dynamical law or requiring Lorentz symmetry of the underlying data.

\textbf{Cross-domain behavior.}
A notable property of MinkowskiPE is that the same positional construction can be applied across substantially different spatiotemporal domains without changing its underlying formulation. Molecular dynamics and video prediction differ not only in physical scale, but also in token semantics, spatial structure, and prediction architecture. Nevertheless, both can be represented through the same event-based view of time and space and modeled with the same positional mechanism. This suggests that MinkowskiPE provides a flexible geometric representation for spatiotemporal relations across different Transformer-based prediction settings.

\textbf{Limitations and Outlook.}
MinkowskiPE adopts a fixed geometric bias rather than learning the geometry from data, and different applications may benefit from different geometric inductive biases. Importantly, this choice does not assume that the underlying dynamics obey Lorentz symmetry. If one instead seeks a positional encoding that explicitly respects the spacetime symmetry of a physical system, Galilean symmetry provides a natural candidate for many non-relativistic settings \citep{LevyLeblond1971}. In Appendix~\ref{app:galilean}, we analyze one such construction based on finite-dimensional unitary token-wise transformations and show that exact Galilean invariance imposes a strong structural constraint: the resulting pairwise interaction cannot retain the Galilean-invariant spatial component of the relative group element.
This result does not rule out Galilean positional encodings more generally, but highlights a nontrivial obstruction for this particular class of constructions.

A second limitation is that our current construction uses learned scalar projections followed by one-parameter Lorentz transformations. More expressive multi-parameter constructions may capture richer interactions among temporal and spatial directions, but require mutually commuting Lorentz generators and are therefore substantially more constrained. We provide further analysis in Appendix~\ref{app:derivation}. Finally, our experiments focus on molecular and visual forecasting. Extending the framework to embodied dynamics, weather forecasting, and generative models, as well as exploring data-adaptive geometric structures, are natural directions for future work.

\vspace{-1ex}
\section{Conclusion}
\vspace{-1ex}

We introduced MinkowskiPE, a relative positional encoding that represents Transformer tokens as events in joint time--space coordinates. Using Lorentz transformations in query--key feature space, it couples temporal and spatial positional information, with attention scores depending on position only through relative spacetime displacement. The construction requires only a lightweight modification to query--key transformations and applies unchanged across different spatiotemporal tokenizations. Across molecular dynamics and video prediction, MinkowskiPE achieves strong performance across distinct spatiotemporal settings. Together, these results highlight positional geometry as a key design axis for spatiotemporal Transformers.

\vspace{-1ex}
\subsection*{AI use statement}
\vspace{-1ex}
The authors used ChatGPT to polish the writing and assist in literature discovery and interpretation. CodeX was used for limited code review, debugging, and utility scripts, while the core implementation was developed by the authors. During experimentation, CodeX assisted with monitoring, aggregating, and organizing experimental results. All AI-assisted material was reviewed and verified by the authors, who take full responsibility for the final manuscript, code, and reported results.

\renewcommand*{\bibfont}{\small}
\printbibliography

\appendix

\section{Additional Derivations of MinkowskiPE}
\label{app:derivation}

This appendix develops the group-theoretic structure underlying the construction in \Cref{sec:method}. 
We first formalize how a homomorphic Lorentzian representation yields an exact relative-position interaction. And then we characterize general smooth homomorphisms from the additive spatiotemporal coordinate group into a Lorentz group, and show that for the $(1+1)$-dimensional feature blocks used by MinkowskiPE, the construction necessarily reduces to a linear scalar projection followed by a one-parameter Lorentz boost.

\subsection{Relative Lorentzian Construction}

Let $x\in\mathbb{R}^{D}$ denote an event coordinate, and let $ \Gamma:(\mathbb{R}^{D},+) \to SO^{+}(1,d_s)$ denote a position-dependent Lorentz representation acting on a $(1+d_s)$-dimensional feature space. We write
\begin{equation}
    \eta_s = \mathrm{diag}(-1,1,\ldots,1)
\end{equation}
for the Minkowski metric on this feature space. By definition, 
\begin{equation}
    \Gamma(x)^\top \eta_s \Gamma(x) = \eta_s,
\end{equation}
and hence
\begin{equation} \label{eq:app_lorentz_inverse}
    \Gamma(x)^\top \eta_s
    = \eta_s \Gamma(x)^{-1}.
\end{equation}

For query and key features $q_i$ and $k_j$ associated with event
coordinates $x_i$ and $x_j$, define
\begin{equation}
    \tilde q_i=\Gamma(x_i)q_i,
    \quad
    \tilde k_j=\eta_s\Gamma(x_j)k_j.
\end{equation}
Their Euclidean dot product becomes
\begin{equation} \label{eq:app_pairwise}
    \tilde q_i^\top\tilde k_j
    = q_i^\top \Gamma(x_i)^\top \eta_s \Gamma(x_j)
    k_j 
    = q_i^\top \eta_s \Gamma(x_i)^{-1} \Gamma(x_j) k_j,
\end{equation}
where the second equality follows from \Cref{eq:app_lorentz_inverse}.

Suppose that $\Gamma$ is a homomorphism from the additive coordinate group into the Lorentz group,
\begin{equation} \label{eq:group_homomorphism}
    \Gamma(x_1 + x_2) = \Gamma(x_1) \Gamma(x_2),
    \quad \Gamma(0)=I.
\end{equation}
Then $\Gamma(-x)=\Gamma(x)^{-1}$, and therefore
\begin{equation}
    \Gamma(x_i)^{-1} \Gamma(x_j)
    = \Gamma(x_j-x_i)
    = \Gamma(\Delta x_{ij}).
\end{equation}
Substituting this relation into \Cref{eq:app_pairwise} gives
\begin{equation} \label{eq:app_relative}
    \tilde q_i^\top\tilde k_j
    = q_i^\top \eta_s \Gamma(\Delta x_{ij}) k_j,
\end{equation}
so the positional contribution depends on the event coordinates only through their relative displacement.
The metric factor $\eta_s$ in the key transformation is essential to this construction. If both query and key were transformed simply as $\Gamma(x_i)q_i$ and $\Gamma(x_j)k_j$, their interaction would instead contain $\Gamma(x_i)^\top\Gamma(x_j)$. Lorentz transformations are generally not orthogonal under the ordinary Euclidean inner product, so $\Gamma(x)^\top\neq\Gamma(x)^{-1}$. The insertion of $\eta_s$ converts the transpose into the group inverse through \Cref{eq:app_lorentz_inverse}, allowing the two absolute transformations to reduce exactly to a transformation of the relative coordinate.

\subsection{Group-Theoretic Characterization}

We next characterize smooth Lorentz representations satisfying
\Cref{eq:group_homomorphism}. Let $ \Gamma:(\mathbb{R}^{D},+) \to SO^{+}(1,d_s)$ be a smooth homomorphism, where $D$ is the dimension of the event coordinate and $1+d_s$ is the dimension of the Lorentzian feature representation.

For each coordinate direction, define the infinitesimal generator
\begin{equation}
    G_\mu = \left. \frac{\partial}{\partial x_\mu}\Gamma(x) \right|_{x=0},
    \quad \mu=1,\ldots,D.
\end{equation}
Since every $\Gamma(x)$ preserves $\eta_s$, each generator belongs to the Lorentz Lie algebra and satisfies
\begin{equation} \label{eq:generator_relation}
    G_\mu^\top \eta_s + \eta_s G_\mu = 0.
\end{equation}

Because $(\mathbb{R}^{D},+)$ is Abelian, the images of different
coordinate directions commute
\begin{equation}
    \Gamma(se_\mu) \Gamma(te_\nu) =
    \Gamma(te_\nu) \Gamma(se_\mu)
\end{equation}
for arbitrary $s,t\in\mathbb{R}$. Differentiating at the identity gives
\begin{equation} \label{eq:commutation}
    [G_\mu, G_\nu] = 0 \quad \text{for all } \mu,\nu.  
\end{equation}
Each coordinate direction therefore generates a one-parameter subgroup,
\begin{equation}
    \Gamma(se_\mu) = \exp(sG_\mu).
\end{equation}
Since $x=\sum_{\mu=1}^{D}x_\mu e_\mu$, the homomorphism property gives
\begin{equation}
    \Gamma(x) = \prod_{\mu=1}^{D} \exp(x_\mu G_\mu).
\end{equation}
Using \Cref{eq:commutation}, this becomes
\begin{equation} \label{eq:exponential_generator}
    \Gamma(x) =
    \exp\left(
        \sum_{\mu=1}^{D}x_\mu G_\mu
    \right).
\end{equation}

Conversely, any collection of mutually commuting Lorentz generators $\{G_\mu\}_{\mu=1}^{D}$ defines a smooth homomorphism through \Cref{eq:exponential_generator}. Thus, within this homomorphic construction, an exact multidimensional relative Lorentzian representation is characterized by a mutually commuting set of Lorentz generators.

\paragraph{The (1+1)-dimensional Case.}
MinkowskiPE operates on two-dimensional Lorentzian feature blocks. For such a block, the relevant connected Lorentz group is $SO^{+}(1,1)$, whose Lie algebra $\mathfrak{so}(1,1)$ is one-dimensional. A basis generator is
\begin{equation}
    G=
    \begin{bmatrix}
        0 & 1\\
        1 & 0
    \end{bmatrix},
    \quad
    \eta_{1,1}
    =
    \begin{bmatrix}
        -1 & 0\\
        0 & 1
    \end{bmatrix},
\end{equation}
which satisfies \Cref{eq:generator_relation}.

Because $\mathfrak{so}(1,1)$ is one-dimensional, every generator $G_\mu$ must be proportional to $G$, \ie{},
\begin{equation}
    G_\mu=w_\mu G
\end{equation}
for some scalar coefficient $w_\mu$. Substituting this relation into \Cref{eq:exponential_generator} gives
\begin{equation} \label{eq:exponential_generator_2}
    \Gamma(x)
    =
    \exp\left(
        \sum_{\mu=1}^{D}x_\mu w_\mu G
    \right)
    =
    \exp\left[
        (w^\top x)G
    \right].
\end{equation}
Defining $\rho=w^\top x$, we therefore obtain
\begin{equation} \label{eq:projection_map}
    \Gamma(x) = \Lambda(\rho) = \Lambda(w^\top x),
\end{equation}
where $\Lambda$ denotes the corresponding one-parameter Lorentz boost.

Since $G^2=I$,
\begin{equation}
    \Lambda(\rho)
    = e^{\rho G}
    = \cosh(\rho)I+\sinh(\rho)G
    =
    \begin{bmatrix}
        \cosh\rho & \sinh\rho\\
        \sinh\rho & \cosh\rho
    \end{bmatrix}.
\end{equation}
Equation~\eqref{eq:projection_map} gives a structural characterization of the scalar-projection form used in MinkowskiPE. Within a (1+1)-dimensional Lorentzian feature block, any smooth homomorphism from the additive event-coordinate group into $SO^{+}(1,1)$ is of the form
$$
    x ~ \longmapsto ~ 
    \rho=w^\top x ~ \longmapsto ~ 
    \Lambda(\rho),
$$
including the trivial case $w=0$. Thus, the learned scalar projection in MinkowskiPE is not merely an implementation convenience but the general smooth homomorphic parameterization within this feature-block class.

\paragraph{Higher-dimensional Lorentz Representations.}
The one-parameter construction above is not the only possible homomorphic Lorentz representation. In higher-dimensional feature spaces, genuinely multi-parameter constructions can be obtained from higher-dimensional commuting subalgebras of $\mathfrak{so}(1,d_s)$. However, the commutation requirement places strong structural constraints on the allowed generators.

For example, let $K_a$ denote the Lorentz boost generator along the $a$-th positive-signature direction and $J_{ab}$ the generator of a rotation in the $(a,b)$ plane. For $a\neq b$,
\begin{equation}
    [K_a,K_b]=-J_{ab},
\end{equation}
up to convention-dependent signs. Hence independent boosts along different directions do not commute and cannot directly serve as independent generators of a homomorphism from $(\mathbb{R}^{D},+)$.

More generally, for noncommuting generators $A$ and $B$, the Baker--Campbell--Hausdorff formula gives
\begin{equation}
    e^Ae^B
    =
    \exp\left(
        A+B+\frac{1}{2}[A,B]+\cdots
    \right),
\end{equation}
so composition introduces additional commutator terms and the simple additive relation
\begin{equation}
    \Gamma(x_1)\Gamma(x_2)
    =
    \Gamma(x_1+x_2)
\end{equation}
does not generally hold.

Higher-dimensional commuting constructions remain possible, including subgroups that act only within positive-signature directions. Our choice of $(1+1)$-dimensional Lorentzian blocks instead gives the minimal feature-space realization that mixes the two metric signatures while admitting an exact one-parameter homomorphic form. More expressive multi-parameter Lorentzian positional encodings would require additional choices of commuting subalgebras and are left for future work.

\section{Additional Experimental Details}
\label{app:experiments}

\subsection{Implementation Details of MinkowskiPE}
\label{app:exp-implementation}

\textbf{MinkowskiPE Implementation.}
In all experiments, MinkowskiPE is applied to the query and key features, while the value features remain unchanged. For each attention head, adjacent feature dimensions are grouped into two-dimensional blocks and transformed using the Lorentzian construction described in \Cref{sec:method}. The same projection matrix $W$ is shared across all Transformer layers and attention heads. For molecular dynamics, the hidden dimension is $128$ with $8$ attention heads, giving a head dimension of $16$ and $P=8$ Lorentz blocks. For video prediction, the hidden dimension is $144$ with $8$ attention heads, giving a head dimension of $18$ and $P=9$ blocks.

\textbf{Coordinate Normalization.}
Before applying MinkowskiPE, event coordinates are normalized to task-dependent ranges. For molecular dynamics, all spatial coordinates within an input window are first centered by the mass-weighted ligand center of mass (COM) at the last observed frame. Let
\begin{equation}
    r_{\max} = \max_{\mathbf r\in\mathcal{R}}
    \left\| \mathbf r-\mathbf r_{\mathrm{COM}}^{\mathrm{last}} \right\|_2 ,
\end{equation}
where $\mathcal{R}$ contains all valid ligand positions in the window and the alpha-carbon (C$\alpha$) positions of the cropped protein residues. The spatial coordinates used by MinkowskiPE are then
\begin{equation}
    \mathbf r_{\mathrm{PE}}
    = \frac{
    \mathbf r-\mathbf r_{\mathrm{COM}}^{\mathrm{last}}
    }{
    \max(10,r_{\max}/2)
    },
\end{equation}
while the temporal coordinates over the observed frames are linearly mapped to $[-1,1]$. For video, the temporal and two spatial coordinates are directly constructed on a regular grid in $[-1,1]^3$. This normalization keeps the inputs in a numerically well-behaved regime for the hyperbolic functions.

\textbf{Numerical Implementation.}
The coordinate projection and the evaluation of $\cosh$ and $\sinh$ are performed in float32, after which the transformed query and key features are cast back to the attention dtype. We implement attention using PyTorch scaled dot-product attention. On the NVIDIA H200 GPUs used in our experiments, this operation is dispatched to the FlashAttention backend.

\subsection{Molecular Dynamics Prediction}
\label{app:exp-md}

\begin{figure}[t]
    \centering
    \includegraphics[width=\linewidth]{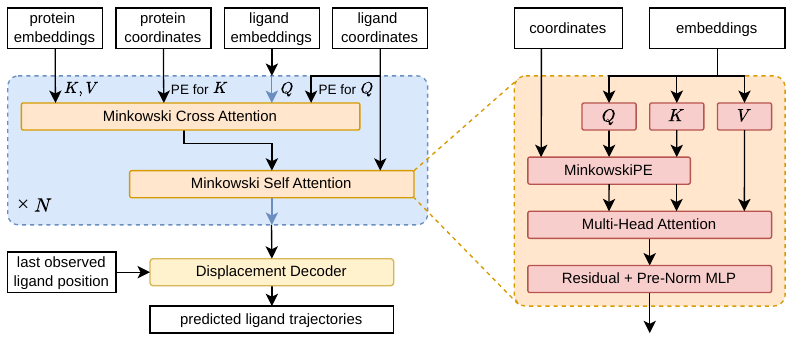}
    \vspace{-0.4cm}
    \caption{%
    Architecture used for molecular dynamics prediction.
    Each block performs frame-wise cross-attention from ligand atoms to the surrounding protein residues, followed by global self-attention over ligand event tokens across the observed trajectory.
    MinkowskiPE is applied to the query--key interactions using the corresponding spatiotemporal event coordinates.
    The representations at the last observed ligand frame are decoded into future atomic displacements.
    }
    \label{fig:pipeline-MD}
\end{figure}

\textbf{Data and Preprocessing.}
We use protein--ligand trajectories from MISATO~\citep{MISATO} and follow the preprocessing pipeline of NeuralMD~\citep{NeuralMD}. Each preprocessed trajectory contains $100$ stored frames. Ligand inputs contain heavy atoms only, whereas the protein is represented at the residue level using backbone N, C$\alpha$, and C atoms. For each input window, we retain at most $128$ protein residues whose C$\alpha$ atoms are closest to the mass-weighted ligand center of mass at the last observed frame. The same center is used to translate the observed ligand, target ligand, and protein coordinates.

Ligand content features include atomic number, atomic mass, centered coordinates, normalized time, frame-to-frame velocity, and a ligand-type embedding. Protein content features include residue identity, the local backbone vectors N--C$\alpha$ and C--C$\alpha$, the centered C$\alpha$ coordinate, and a protein-type embedding. No geometric data augmentation is applied. Although the protein structure is static, its positional coordinate in frame-wise cross-attention is paired with the temporal coordinate of the corresponding ligand frame. Specifically, a protein residue with C$\alpha$ position $(x_j, y_j, z_j)$ is assigned the event coordinate $(t_i, x_j, y_j, z_j)$ when attending to ligand atoms at frame $t_i$.

\textbf{Single- and Multi-trajectory Settings.}
For single-trajectory prediction, we follow the NeuralMD protocol and train a separate model for each of the ten protein--ligand systems reported in \Cref{tab:md_single}. Training windows contain $10$ observed frames and $20$ target frames, following the same in10/out20 split and protocol as NeuralMD. For multi-trajectory prediction, we use MISATO-100, MISATO-1000, and MISATO-Full. After preprocessing, the three datasets contain $80/10/10$, $800/100/100$, and $13{,}066/ 1{,}357/ 1{,}357$ training/validation/test complexes, respectively. Each trajectory contributes one window containing $80$ observed frames followed by $20$ target frames.

\textbf{Model Architecture.}
The molecular model consists of four repeated blocks, each containing frame-wise ligand-to-protein cross-attention followed by global self-attention over all ligand event tokens in the observed trajectory, as illustrated in \Cref{fig:pipeline-MD}. The hidden dimension is $128$, with $8$ attention heads and a $512$-dimensional feed-forward layer. Both the attention and feed-forward sublayers use pre-normalization and residual connections, with dropout $0.1$. A final LayerNorm is applied after the Transformer stack.

After the final block, only the representation of each ligand atom at the last observed frame is passed to the prediction head. The displacement decoder consists of $128 \to 512 \to 60$ with a GELU nonlinearity and dropout $0.1$. The $60$ outputs correspond to the three-dimensional displacements of the $20$ future frames. All future frames are predicted in parallel, and the predicted displacements are added to the last observed ligand positions.

\textbf{Optimization Objective.}
The model is trained using a combination of coordinate prediction error and pairwise structural consistency. For a predicted ligand trajectory $\hat{\mathbf r}_{f,i}$ and target trajectory $\mathbf r_{f,i}$, we use
\begin{equation}
    \mathcal{L}
    = \mathcal{L}_{\mathrm{coord}}
    + 2 \mathcal{L}_{\mathrm{pair}},
\quad 
\text{where} 
\quad 
    \mathcal{L}_{\mathrm{coord}}
    = \frac{
    \sum_{f,i} \left\| \hat{\mathbf r}_{f,i} - \mathbf r_{f,i} \right\|_1
    }{
    \sum_f N_f
    },
\end{equation}
and $\mathcal{L}_{\mathrm{pair}}$ is the mean SmoothL1 loss, with $\beta=0.5$, between predicted and target inter-atomic distances over all valid unordered atom pairs $i<j$ and future frames.

\textbf{Training Details.}
We train all molecular dynamics models with Adam, zero weight decay, and a linear warmup followed by cosine learning-rate decay. The backbone is warmed up for two epochs, while the decoder learning rate is applied without warmup.
For the single-trajectory setting, both the backbone and decoder use a learning rate of $10^{-4}$, and four temporal windows are accumulated per optimizer step. 
For the multi-trajectory setting, the backbone learning rate is $3\times10^{-5}$, the decoder learning rate remains $10^{-4}$, and the batch size is eight complexes. 
Training uses BF16 mixed precision for at most $100$ epochs with early-stopping patience $20$. No gradient clipping is applied. All reported results use random seeds $0$, $42$, and $123$.

\textbf{Evaluation Metrics.}
We follow the evaluation definitions of NeuralMD. Let
$e_{f,ij} = \hat d_{f,ij} - d_{f,ij}$
denote the error between predicted and target pairwise distances at future frame $f$. The coordinate error is measured by
\begin{equation}
    \mathrm{MAE} = \frac{
    \sum_{f,i} \left\| \hat{\mathbf r}_{f,i} - \mathbf r_{f,i} \right\|_1
    }{
    \sum_f N_f
    },
\end{equation}
where the $\ell_1$ norm sums the errors over the three Cartesian coordinates.

For each evaluated sample and frame, Matching is the root-mean-square error over all valid ordered atom pairs,
\begin{equation}
    \mathrm{Matching}_f = \sqrt{\frac{1}{|\mathcal{P}_f|} \sum_{(i,j)\in\mathcal{P}_f} e_{f,ij}^{\,2}},
\end{equation}
where $\mathcal{P}_f$ includes the diagonal and both $(i,j)$ and $(j,i)$. The reported Matching score is averaged over all evaluated sample--frame pairs.

Stability measures the fraction of pairwise distances whose absolute error does not exceed $0.5$~\AA:
\begin{equation}
    \mathrm{Stability}_f = 100 \times 
    \frac{
    \sum_{(i,j)\in\mathcal{P}_f} \mathbbm{1}\!\left[ |e_{f,ij}|\leq 0.5~\text{\AA}\right]
    }{
    |\mathcal{P}_f|
    }.
\end{equation}
The final Stability score is the average over all evaluated sample--frame pairs.

For the single-trajectory comparison in \Cref{tab:md_single}, the VerletMD, GNN-MD, NeuralMD-ODE, NeuralMD-SDE, and DenoisingLD results are taken from NeuralMD, while BioMD and MinkowskiPE are evaluated in our implementation under the same data split and evaluation protocol.
For the multi-trajectory comparison in \Cref{tab:md_multiple}, all baselines are retrained under the in80/out20 protocol described above.

\subsection{Video Prediction}
\label{app:exp-video}

\textbf{Data and Preprocessing.}
We follow the OpenSTL~\citep{OpenSTL} preprocessing and evaluation protocol for the KTH human-action dataset \citep{KTH}. Persons 01--16 are used for training and persons 17--25 for testing, producing $5{,}200$ training clips and $3{,}167$ test clips. Each clip contains $30$ consecutive grayscale frames, with the first $10$ used as observations and the subsequent $20$ as prediction targets. All frames are resized to $128\times128$ and pixel intensities are scaled to $[0,1]$.

During training, the same spatial augmentation is applied synchronously to all frames in a clip. The frames are first bilinearly resized by a factor of $1/0.95$, followed by a random $128\times128$ crop and horizontal flipping with probability $0.5$. Following the OpenSTL preprocessing protocol, 30-frame clips are extracted with start-index strides of 3 frames for jogging and running, and 30 frames for boxing, handclapping, handwaving, and walking.

\textbf{Model Architecture.}
Each observed grayscale frame is concatenated with its frame difference, giving a two-channel input. The convolutional encoder consists of three stages, $2 \to 36 \to 72 \to 144$, reducing the spatial resolution from $128\times128$ to $16\times16$. Each stage uses a $4\times4$ convolution with stride $2$, followed by GroupNorm and GELU. The resulting latent feature dimension is $144$.

The $10$ observed latent maps are combined with $20$ future latent maps, initialized from the final observed latent map. Each future token is augmented with time-step and token-type embeddings, together with a velocity embedding computed from the difference between the final two observed latent maps. The velocity embedding is modulated by a zero-initialized learned gate.
The resulting $30\times16\times16=7{,}680$ event tokens are jointly processed by an eight-layer global Transformer equipped with MinkowskiPE. The Transformer has hidden dimension $144$, $8$ attention heads, and a $576$-dimensional feed-forward layer.

The final $20$ latent maps are decoded through progressive bilinear upsampling from $16\times16$ to $32\times32$, $64\times64$, and $128\times128$. The first two decoding stages use skip connections from the corresponding encoder features of the last observed frame. The convolutional channel dimensions of the three stages are
\begin{equation*}
    216\to 72\to 72, \qquad
    108\to 36\to 36, \qquad
    36\to 18\to 18,
\end{equation*}
followed by a final $3\times3$ convolution from $18$ channels to one output channel. The decoder predicts residuals relative to the last observed frame, and all $20$ future frames are generated in parallel.

The resulting model contains $2{,}482{,}086$ parameters in total, reported as $2.5$M in \Cref{tab:video}.

\textbf{Training and Evaluation.}
The model is trained using the mean squared error between predicted and target frames. We use Adam with a learning rate of $10^{-3}$, zero weight decay, and a OneCycle learning-rate schedule. The global batch size is $16$, the gradient norm is clipped at $1.0$, and training uses BF16 mixed precision. We train for at most $100$ epochs with an early-stopping patience of $20$ epochs. We follow the OpenSTL evaluation protocol. Within each run, we select the checkpoint with the lowest MSE on the test dataset. We report MSE, MAE, SSIM, PSNR, and LPIPS following the metric definitions used in OpenSTL, with LPIPS computed using the AlexNet backbone. All experiments use random seeds 0, 42, and 123.

\section{Constraints on Galilean-Invariant Positional Encoding}
\label{app:galilean}

Our use of Lorentz transformations provides a representation-level geometric bias and does not impose Lorentz symmetry on the underlying dynamics. If one instead seeks a positional encoding that respects the physical symmetry of non-relativistic dynamics, Galilean symmetry is a natural starting point \citep{LevyLeblond1971}.

This appendix examines a particular construction based on finite-dimensional unitary token-wise transformations. We show that exact invariance under common Galilean transformations forces the resulting pairwise interaction to discard the spatial component of the relative group element. This obstruction depends on the unitary assumption, which is not satisfied by the Lorentz boosts used in MinkowskiPE. It therefore identifies a constraint on this specific route to Galilean-invariant positional encoding.

\textbf{Galilean Relative Structure.}
Consider first the rotation-free Galilei group $G_0$, generated by spatial translations, time translations, and Galilean boosts. An element is written as
\begin{equation}
    g=(a,b,v),
\end{equation}
where $a\in\mathbb{R}^d$ is a spatial translation, $b\in\mathbb{R}$ is a time translation, and $v\in\mathbb{R}^d$ is a boost. The group law is
\begin{equation} \label{eq:galilean_group_law}
    (a_1, b_1, v_1) (a_2, b_2, v_2)
    = (a_1+a_2+v_1 b_2, \, b_1+b_2,\, v_1+v_2).
\end{equation}
We denote the corresponding subgroups by
\begin{equation} \label{eq:galilean_subgroups}
    T(a) = (a,0,0), \quad
    S(b) = (0,b,0), \quad
    B(v) = (0,0,v).
\end{equation}

For spacetime coordinates $(x,t)$ alone, a common Galilean boost acts as
\begin{equation}
    x \mapsto x + vt, \quad t \mapsto t,
\end{equation}
and therefore transforms a relative displacement according to
\begin{equation}
    \Delta x_{ij} \mapsto \Delta x_{ij} + v \Delta t_{ij}.
\end{equation}
Thus, unlike in the translational setting used by MinkowskiPE, $\Delta x_{ij}$ itself is not invariant under a common Galilean boost.

To obtain a boost-invariant relative spatial quantity, one may augment each token with a velocity $u$ and write
\begin{equation}
    z = (x, t, u).
\end{equation}
Identifying such a state with the group element
\begin{equation}
    g(z) = (x, t, u),
\end{equation}
the relative group element is
\begin{equation} \label{eq:galilean_relative_element}
    g(z_i)^{-1} g(z_j) = (r_{ij}, \Delta t_{ij}, \Delta u_{ij}),
\end{equation}
where
\begin{equation}
    r_{ij} = x_j - x_i - u_i (t_j - t_i),
    \quad
    \Delta t_{ij} = t_j - t_i,
    \quad
    \Delta u_{ij} = u_j - u_i.
\end{equation}
Under a common Galilean boost,
\begin{equation}
    x \mapsto x + vt, \quad u \mapsto u + v,
\end{equation}
the quantity $r_{ij}$ remains unchanged. Hence $r_{ij}$ is the natural spatial component of the relative Galilean group element.

\textbf{A Finite-dimensional Obstruction.}
We now consider a direct rotary-style token-wise encoding.
Let $U: G_0 \to U(M)$ be a continuous finite-dimensional unitary encoding, where $M$ is the feature dimension, and define the pairwise operator
\begin{equation} \label{eq:galilean_pairwise_operator}
    A(g_i, g_j) = U(g_i)^\dagger U(g_j).
\end{equation}
We require invariance under a common Galilean transformation,
\begin{equation} \label{eq:galilean_pairwise_invariance}
    A(hg_i, hg_j) = A(g_i,g_j)
    \quad \text{for all } h, g_i, g_j \in G_0.
\end{equation}
This is the direct group-theoretic analogue of the relative-position property underlying rotary positional encodings.

\textbf{Proposition 1.}
\textit{
Let $U: G_0 \to U(M)$ be continuous, with $M$ finite, and suppose that the pairwise operator in \Cref{eq:galilean_pairwise_operator} satisfies
\Cref{eq:galilean_pairwise_invariance}. Then $A(g_i, g_j)$ is independent of the spatial component $r_{ij}$ of the relative group element $g_i^{-1} g_j$. It may depend only on $\Delta t_{ij}$ and $\Delta u_{ij}$.
}

\textit{Proof.}
Define
\begin{equation}
    F(g)=A(e,g).
\end{equation}
Choosing $h=g_i^{-1}$ in \Cref{eq:galilean_pairwise_invariance} gives
\begin{equation} \label{eq:proof_1}
    A(g_i, g_j) = F(g_i^{-1} g_j).
\end{equation}
Moreover, the token-wise factorization in \Cref{eq:galilean_pairwise_operator} implies
\begin{equation} \label{eq:proof_2}
    A(g_i, g_j) A(g_j, g_k) = A(g_i, g_k).
\end{equation}
Combining \Cref{eq:proof_1} and \Cref{eq:proof_2}, and writing $a = g_i^{-1} g_j$, $c = g_j^{-1} g_k$, we obtain
\begin{equation}
    F(a) F(c) = F(ac).
\end{equation}
Thus $F$ is a continuous finite-dimensional unitary representation of $G_0$.

Let $P_r$, $H$, and $K_r$ denote the Hermitian generators of spatial translations, time translations, and boosts:
\begin{equation}
    F(T(a)) = e^{-i a\cdot P}, \quad
    F(S(b)) = e^{-ibH}, \quad
    F(B(v)) = e^{-iv \cdot K}.
\end{equation}
The Galilei group law in \Cref{eq:galilean_group_law} gives
\begin{equation}
    B(v) S(b) B(v)^{-1} S(b)^{-1} = T(vb),
\end{equation}
and therefore the generators satisfy
\begin{equation} \label{eq:boost_time_commutator}
    [K_r, H] = i P_r.
\end{equation}
Spatial translations are central in the rotation-free subgroup $G_0$, so
\begin{equation} \label{eq:translation_central}
    [P_r, K_s] = 0, \qquad [P_r, H] = 0.
\end{equation}

Using \Cref{eq:boost_time_commutator} and \Cref{eq:translation_central},
\begin{equation}
    \mathrm{tr}(P_r^2)
    = -i \,\mathrm{tr}\! \left( P_r [K_r, H] \right)
    = -i \,\mathrm{tr}\! \left( [P_r K_r, H] \right)
    = 0.
\end{equation}
Because $P_r$ is Hermitian, $\mathrm{tr}(P_r^2)$ is the sum of the squared eigenvalues of $P_r$. Hence $P_r = 0$ for every $r=1, \cdots, d$. It follows that
\begin{equation}
    F(T(r)) = I \quad \text{for all } r \in \mathbb{R}^d.
\end{equation}

Using \Cref{eq:galilean_relative_element} and \Cref{eq:proof_1}
\begin{equation}
    A(g_i, g_j) 
    = F\!\left( T(r_{ij}) S(\Delta t_{ij}) B(\Delta u_{ij}) \right)
    = F\!\left( S(\Delta t_{ij}) B(\Delta u_{ij}) \right),
\end{equation}
which is independent of $r_{ij}$.
\hfill$\square$ \vspace{0.5em}

\textbf{Implications for Positional Encoding.}
The proposition does not imply that Galilean positional encoding is impossible in general. Rather, it identifies an obstruction specific to exact, finite-dimensional, unitary, token-wise factorizations of the form in \Cref{eq:galilean_pairwise_operator}. Such an encoding may still represent relative time and relative velocity. For example, one may define
\begin{equation}
    U(z) = \bigoplus_{k=1}^{M/2} R\!\left( \omega_k t+\kappa_k^\top u \right),
\end{equation}
where $R(\cdot)$ is a planar rotation. Its pairwise interaction depends on
\begin{equation}
    \omega_k \Delta t_{ij} + \kappa_k^\top \Delta u_{ij},
\end{equation}
but contains no dependence on the Galilean-invariant spatial quantity $r_{ij}$, in agreement with Proposition~1.

Other approaches, including explicitly pairwise constructions based on Galilean invariants, non-unitary representations, and infinite-dimensional representations, fall outside the assumptions of Proposition 1. Projective representations and central extensions provide further directions that require a separate analysis \citep{Bargmann1954,LevyLeblond1971}. These alternatives remain open possibilities for incorporating Galilean structure into positional encoding.

\end{document}